**Title:** Associations Between Natural Language Processing (NLP) Enriched Social Determinants of Health and Suicide Death among US Veterans

**Authors**:
Avijit Mitra, MSc[1]
Richeek Pradhan, MD[2]
Rachel D Melamed, PhD[3]
Kun Chen, PhD[4,5]
David C Hoaglin, PhD[6]
Katherine L Tucker, PhD[7]
Joel I Reisman, AB[8]
Zhichao Yang, MS[1]
Weisong Liu, PhD[9,10]
Jack Tsai, PhD[11,12]
Hong Yu, PhD [1,8,9,10]

[1] Manning College of Information and Computer Sciences, University of Massachusetts Amherst, MA, USA
[2] Department of Epidemiology, Biostatistics and Occupational Health, McGill University, Montreal, Quebec, Canada
[3] Department of Biological Sciences, University of Massachusetts Lowell, Lowell, MA, USA
[4] Department of Statistics, University of Connecticut, Storrs, CT, USA
[5] Center for Population Health, Uconn Health, Farmington, CT, USA
[6] Department of Population and Quantitative Health Sciences, University of Massachusetts Chan Medical School, Worcester, MA, USA.
[7] Department of Biomedical & Nutritional Sciences, University of Massachusetts Lowell, Lowell, MA, USA
[8] Center for Healthcare Organization & Implementation Research, Veterans Affairs Bedford Healthcare System, Bedford, MA, USA
[9] Miner School of Computer & Information Sciences, University of Massachusetts Lowell, Lowell, MA, USA
[10] Center for Biomedical and Health Research in Data Sciences, University of Massachusetts Lowell, Lowell, MA, USA
[11] National Center on Homelessness Among Veterans, United States Department of Veterans Affairs, Tampa, FL, USA
[12] School of Public Health, University of Texas Health Science Center at Houston, Houston, TX, USA

**Corresponding Author:**
Hong Yu, PhD
Department of Computer Science,
University of Massachusetts Lowell,
1 University Avenue
Lowell, MA, US
Phone: 1 508 612 7292
Email: Hong_Yu@uml.edu

Date of Revision: 12/06/2022
**Manuscript Word Count:** 2,975

**Abstract**

**Importance:** Social determinants of health (SDOH) are known to be associated with increased risk of suicidal behaviors, but few studies utilized SDOH from unstructured electronic health record (EHR) notes.

**Objective:** To investigate associations between suicide and recent SDOH, identified using structured and unstructured data.

**Design:** Nested case-control study.

**Setting:** EHR data from the US Veterans Health Administration (VHA).

**Participants:** 6,122,785 Veterans who received care in the US VHA between October 1, 2010, and September 30, 2015.

**Exposures:** Occurrence of SDOH over a maximum span of two years compared with no occurrence of SDOH.

**Main Outcomes and Measures:** Cases of suicide deaths were matched with 4 controls on birth year, cohort entry date, sex, and duration of follow-up. We developed an NLP system to extract SDOH from unstructured notes. Structured data, NLP on unstructured data, and combining them yielded six, eight and nine SDOH respectively. Adjusted odds ratios (aORs) and 95% confidence intervals (CIs) were estimated using conditional logistic regression.

**Results:** In our cohort, 8,821 Veterans committed suicide during 23,725,382 person-years of follow-up (incidence rate 37.18 /100,000 person-years). Our cohort was mostly male (92.23%) and white (76.99%). Across the five common SDOH as covariates, NLP-extracted SDOH, on average, covered 80.03% of all SDOH occurrences. All SDOH, measured by structured data and NLP, were significantly associated with increased risk of suicide. The SDOH with the largest effects was legal problems (aOR=2.66, 95% CI=2.46-2.89), followed by violence (aOR=2.12, 95% CI=1.98-2.27). NLP-extracted and structured SDOH were also associated with suicide.

**Conclusions and Relevance:** NLP-extracted SDOH were always significantly associated with increased risk of suicide among Veterans, suggesting the potential of NLP in public health studies.

**Introduction**

Suicide is one of the leading causes of death among US residents, accounting for 47,511 deaths in 2019 alone[1]. Nationwide, deaths by suicide increased 30% from 1999 to 2016[2]. In 2013 alone, the total cost of suicides and suicide attempts in the US was estimated to be $93.5 billion[3]. In the past decade, suicide rates have been consistently higher among Veterans than nonveterans, and even more alarming, the suicide rate among Veterans has risen faster than among nonveteran adults[4].

Social determinants of health (SDOH), which include conditions such as socioeconomic status, access to healthy food, education, housing, and physical environment[5], are strong predictors of suicidal behaviors (ideation, attempt and death)[6–9]. For example, social disruptions (e.g., relationship dissolution, financial insecurity, legal problems, or exposure to childhood adversity) are well-known to instigate suicidal behavior[6,10–12]. To formulate policies addressing suicide prevention, one must go beyond identifying predictors by determining the magnitude of the effects of SDOH on suicide. A key impediment to this has been the lack of comprehensive and reliably available SDOH information in large population-based databases, where investigators have traditionally relied on structured data. Structured data often lack completeness regarding SDOH information, specifically, when they are designed for billing purposes. A recent study showed that unstructured data contain about 90 times more SDOH information than structured data[13].

Though existing studies identified a range of common risk-factors for suicide using structured data from electronic health records (EHR)[14–18], unstructured EHR data received little attention in investigating potential links between suicide and SDOH. Therefore, in a nested case-control study, we used both structured data (ICD codes, stop codes) and unstructured data (clinical notes, processed by a novel natural language processing (NLP) system) from the large EHR system of the US Veterans Health Administration (VHA) to examine the association of 9 SDOH factors with risk of suicide.

## Methods

### Data Source

This study used the EHR database from the VHA Corporate Data Warehouse (CDW). With a primary obligation to provide medical services to all eligible US Veterans, the VHA is the largest integrated healthcare network in the country; its EHR system spans more than a thousand medical centers and clinics[19]. The VHA database includes patient demographic information, medication, diagnoses, procedures, clinical notes, and billing. Our study protocol was approved by the institutional review board of US Veterans Affairs (VA) Bedford Health Care, and we obtained a waiver of informed consent.

### Study Population

As with any state-of-the-art NLP system, analyzing all patients in our base cohort presented a computational challenge. In addition, we studied multiple exposures. Therefore, we employed a nested case-control design and used risk-set sampling to match the controls to the cases. This approach facilitates studying associations of exposures (e.g., SDOH) on rare events such as suicide outcome[20].

The base cohort consisted of all Veterans for whom the VHA database had any record of service during the period between 10/01/2010 (start of Fiscal Year [FY] 2011) and 09/30/2015 (end of FY2015). Each patient's cohort entry date was defined as the latest of these dates: when the patient had two years of medical history in the database, the patient's 18th birthday, or the start of FY2011. The end of follow-up was defined as the earliest of the following: suicide, death from other causes, end of last record for the patient, or end of the study period (end of FY2015). We excluded all patients who had prior suicide attempts[21] or no EHR notes before their cohort entry dates. Patients with missing or erroneous demographic information and those older than 100 years of age were also excluded.

Cases consisted of all patients in the base cohort who died by suicide (according to National Death Index[22] with International Statistical Classification of Diseases and Related Health Problems, Tenth Revision, codes X60–X84, Y87.0, and/or U03 as underlying cause of death) during FY2011-2015. Each case was randomly matched, with replacement, to 4 controls from those who were still alive. The matching criteria were - 1) birth year (± 3 years), 2) cohort entry FY, 3) sex and 4) duration of follow-up (same or longer than the case). By design, a case could serve as a control for another case who committed suicide at an earlier date, and a patient could be a control for multiple cases. The index date for each case was defined as the date of suicide and each control was assigned the same index date as their corresponding case. All data analyses were performed in May 2022. The Strengthening the Reporting of Observational Studies in Epidemiology (STROBE)[23] reporting guidelines were followed.

### Natural Language Processing

A unique aspect of this study is the integration of an NLP system to extract SDOH, behavioral and other relevant factors from EHR notes. We implemented a multi-task learning (MTL) framework based on the pre-trained language model - RoBERTa[24]. RoBERTa is an improved version of Bidirectional Encoder Representations from Transformers (BERT)[25] which has been shown to outperform all other NLP systems across a wide range of downstream tasks. To train our MTL model, we collected 4,646 EHR notes from 1,393 patients (excludes all patients from our base cohort) who received treatment at the VHA and died between fiscal year 2009-2017. Under expert supervision, three trained annotators annotated these notes for 13 distinct SDOH, behavioral and relevant factors. This is in accordance with the recent clinical practice guideline issued jointly by VA and Department of Defense[26]. Our MTL model was fine-tuned on this dataset for three joint tasks: factor, presence, and period identification. **Appendix 1** provides a detailed description of our note selection, annotation process and MTL model performance.

We used this fine-tuned MTL model to extract 13 factors for our study population – 8 of which were SDOH (eTable1, **Appendix 1**). For each patient visit, we used 7 types of notes - emergency department notes, nursing assessments, primary care notes, hospital admission notes, inpatient progress notes, pain management, and discharge summaries. For multiple notes within an observation window (covariate or exposure assessment periods), we merged their factor predictions and prioritized presence 'yes' over presence 'no'. Each prediction was dichotomized using the following strategy:
1. Prediction of a factor with presence 'yes' and period 'current' was coded as 1.
2. All other predictions were coded as 0, including factors with missing presence/period attributes.
3. Notes with no factor prediction, and patients with no notes, were coded as 0.

**SDOH Extraction**
We extracted SDOH from both unstructured EHR text (using NLP) and structured data (using ICD codes, and VHA stop codes, details in **Appendix 2**). The NLP-extracted SDOH comprised 8 factors: social isolation, job or financial insecurity, housing instability, legal problems, violence, barriers to care, transition of care, and food insecurity; and the structured SDOH comprised 6 factors[6] - social or familial problems, employment or financial problems, housing instability, legal problems, violence, and non-specific psychosocial needs. We combined these two groups to have 9 distinct factors - 5 were represented in both sets (more in Table 2), whereas barriers to care, transition of care and food insecurity were found only in the NLP-extracted SDOH, and non-specific psychosocial needs, only in the structured SDOH.

**Exposure and Covariate Assessment**
To focus on the impact of recent SDOH events on suicide, we considered a patient's exposures to the aforementioned 9 SDOH in the two years before the index date, but not prior to the cohort entry date. The covariate assessment period was two years prior to the cohort entry date. Potential covariates were socio-demographic variables, clinical comorbidities, and mental health disorders. Socio-demographic variables included race, age, and marital status. From the Charlson comorbidity index[27], we included 17 clinical comorbidities: acute myocardial infarction, congestive heart failure, peripheral vascular disease, cerebrovascular disease, dementia, chronic obstructive pulmonary disease, rheumatoid disease, peptic ulcer disease, mild liver disease, diabetes without complications, diabetes with complications, hemiplegia or paraplegia, renal disease, cancer, moderate or severe liver disease, metastatic solid tumor, and AIDS/HIV. We considered 7 mental health disorders: major depressive disorder, alcohol use disorder, drug use disorder, anxiety disorder, posttraumatic stress disorder, schizophrenia, and bipolar disorder[6]. We also added psychiatric symptoms, substance abuse, pain, and patient disability that were extracted by our NLP system. Additionally, for each model with a specific SDOH as the exposure, the list of covariates included all SDOH in its group (i.e., NLP-extracted, or structured or combined).

**Statistical Analysis**

We calculated the crude incidence rate of suicide along with 95% confidence intervals (CIs) based on the Poisson distribution. For each SDOH exposure variable, we fit a conditional logistic regression model with death by suicide as the outcome. In addition to matching for birth year, cohort entry date, sex, and duration of follow-up, we adjusted all models for the specified covariates (potential confounders). This procedure yielded a total of 23 models: 8 for NLP-extracted SDOH, 6 for structured SDOH, and 9 when combined. Following the same process, we also considered exposure to two SDOH at the same time, yielding a total of 79 models: 28 for NLP-extracted SDOH, 15 for structured SDOH, and 36 when combined. The variance inflation factor (VIF)[28] showed no evidence of collinearity (no VIF exceeded 3) among the covariates for any model. We report adjusted odds ratios (aOR) with 95% CIs for each exposure. All analyses used RStudio (version 0.99.902 with R version 3.6.0).

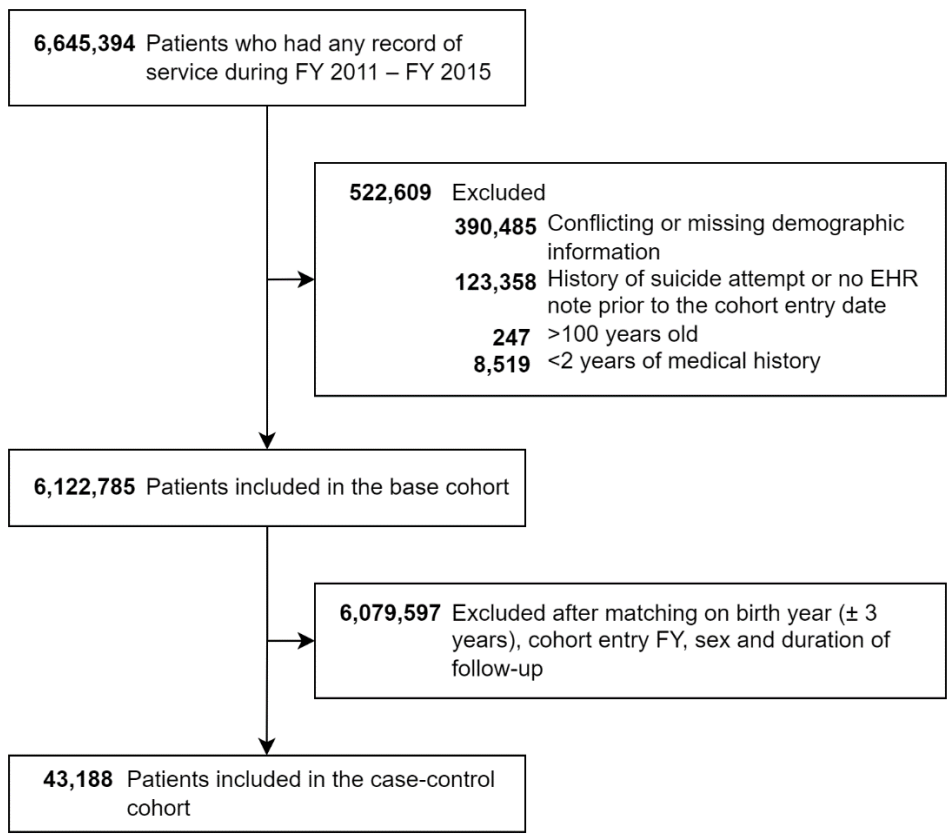

(a)

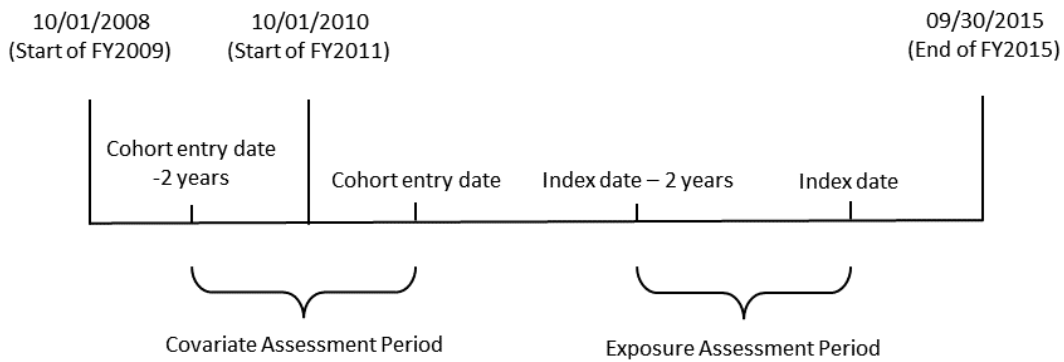

**(b)**
**Figure 1.** (a) Construction of the cohort and (b) our study timeline

## Results
Our base cohort consisted of 6,122,785 Veterans (Figure 1) from 1,185 VA healthcare facilities; the majority were white (76.99%) and male (92.23%). Veterans 50 years of age or older comprised 75.78% of the population. We had a mean follow-up of 3.87 years, generating 23,725,382 person-years, with 8,821 deaths by suicide (crude incidence rate 37.18 per 100,000 person-years, 95% CI = 36.41-37.96). **Appendix 3** reports detailed characteristics of the base cohort. In our case-control cohort, the majority were also white (79.20%) and male (96.45%) with a mean (±SD) age of 58.64±17.41 years. Compared with cases, controls had a higher percentage of black individuals (16.27% vs. 5.50%) and lower percentage of white individuals (76.88% vs. 88.48%). This case-control cohort consisted of 8,821 cases and 35,284 matched controls (34,404 unique individuals, of whom 846 served as controls for more than one case) from 43,188 Veterans who were served at 908 VA facilities. Detailed characteristics are shown in Table 1.

|  | Cases (%) | Controls (%) |
|---|---|---|
| No of Patients | 8,821 | 35,284[a] |
| Race | | |
| Asian | 42 (0.48) | 322 (0.91) |
| American Indian | 68 (0.77) | 240 (0.68) |
| Black | 485 (5.50) | 5,742 (16.27) |
| Native Hawaiian or Other Pacific Islander | 70 (0.79) | 345 (0.98) |
| White | 7,805 (88.48) | 27,125 (76.88) |
| Unknown | 351 (3.98) | 1,510 (4.28) |
| Sex | | |
| Male | 8,508 (96.45) | 34,032 (96.45) |
| Female | 313 (3.55) | 1,252 (3.55) |
| Age | | |
| 18-29 | 774 (8.77) | 3,085 (8.74) |
| 30-39 | 714 (8.09) | 2,842 (8.05) |
| 40-49 | 1,000 (11.34) | 3,911 (11.08) |
| 50-59 | 1,587 (17.99) | 6,163 (17.47) |
| 60-69 | 2,278 (25.82) | 9,597 (27.20) |
| 70-79 | 1,347 (15.27) | 5,271 (14.94) |
| 80-100 | 1,121 (12.71) | 4,415 (12.51) |
| Marital Status | | |
| Married | 2,539 (28.78) | 8,846 (25.07) |
| Single | 1,028 (11.65) | 2,002 (5.67) |
| Divorced | 1,871 (21.21) | 3,117 (8.83) |
| Widowed | 466 (5.28) | 991 (2.81) |
| Unknown | 2,917 (33.06) | 20,328 (57.61) |
| Comorbidities (Charlson) | | |
| Acute Myocardial Infarction | 58 (0.66) | 207 (0.59) |
| Congestive Heart Failure | 213 (2.41) | 812 (2.30) |
| Peripheral Vascular Disease | 281 (3.19) | 948 (2.69) |
| Cerebrovascular disease | 275 (3.12) | 1,081 (3.06) |
| Dementia | 17 (0.19) | 153 (0.43) |
| COPD | 962 (10.91) | 2,859 (8.10) |
| Rheumatoid Disease | 57 (0.65) | 249 (0.71) |
| Peptic Ulcer Disease | 47 (0.53) | 127 (0.36) |
| Mild Liver Disease | 188 (2.13) | 496 (1.41) |
| Diabetes without Complications | 1,108 (12.56) | 5,408 (15.33) |
| Diabetes with Complications | 257 (2.91) | 1246 (3.53) |
| Hemiplegia or Paraplegia | 59 (0.67) | 186 (0.53) |
| Renal Disease | 131 (1.49) | 682 (1.93) |
| Cancer (any malignancy) | 489 (5.54) | 1,673 (4.74) |
| Moderate or Severe Liver Disease | 15 (0.17) | 47 (0.13) |
| Metastatic Solid Tumor | 28 (0.32) | 74 (0.21) |
| AIDS/HIV | 26 (0.29) | 141 (0.40) |
| Comorbidities (Mental Health Disorders) | | |
| Major Depressive disorder | 2,191 (24.84) | 4,790 (13.58) |
| Alcohol Use Disorder | 778 (8.82) | 1,470 (4.17) |
| Drug Use Disorder | 425 (4.82) | 924 (2.62) |
| Anxiety Disorder | 880 (9.98) | 1,953 (5.54) |
| Posttraumatic Stress Disorder | 1,208 (13.69) | 3,676 (10.42) |
| Schizophrenia | 262 (2.97) | 525 (1.49) |
| Bipolar disorder | 708 (8.03) | 1,101 (3.12) |
| NLP-extracted non-SDOH factors | | |
| Patient Disability | 3,718 (42.15) | 13,643 (38.67) |
| Substance Abuse | 5,408 (61.31) | 20,539 (58.21) |
| Psychiatric Symptoms | 5,436 (61.63) | 19,954 (56.55) |
| Pain | 5,664 (64.21) | 22,048 (62.49) |

Abbreviations: COPD, Chronic Obstructive Pulmonary Disease
[a]Number of controls, not number of unique patients.

**Table 1.** Summary Statistics for Cases and Controls

The structured SDOH had low prevalence compared with their NLP-extracted counterparts (Table 2). For example, of all Veterans exposed to 'Social problems', only 32.31% were identified by structured SDOH whereas NLP identified 85.80%. We found similar results for the remaining combined SDOH (**Appendix 4**). As covariates, across all the 5 common SDOH, considering combined SDOH as the gold standard, NLP-extracted SDOH had a coverage of 78.86% on average compared with 36.02% from structured SDOH. As exposures, the numbers were 80.03% and 38.17% respectively. All SDOH occurred more frequently among cases than among controls. Moreover, for the majority of the SDOH, we found more occurrences during the exposure assessment period than during the covariate assessment period.

| | As Covariate | | | As Exposure | | |
|---|---|---|---|---|---|---|
| | Structured Data Case (%)/Control(%) | Combined Case(%)/Control(%) | NLP-extracted Case(%)/Control(%) | Structured Data Case (%)/Control(%) | Combined Case(%)/Control(%) | Combined Case(%)/Control(%) |
| | 833 (9.44%)/ 2,691 (7.63%) | 2,938 (33.31%)/ 9,227 (26.15%) | 3,080 (34.92%)/ 7,809 (22.13%) | 1,291 (14.64%)/ 2,810 (7.96%) | 3,517 (39.87%)/ 9,174 (26.00%) | |
| | 603 (6.84%)/ 1,661 (4.71%) | 2,206 (25.01%)/ 6,652 (18.85%) | 2,274 (25.78%)/ 5,710 (16.18%) | 882 (10.00%)/ 1,858 (5.27%) | 2,513 (28.49%)/ 6,381 (18.08%) | |
| | 592 (6.171%)/ 1,657 (4.70%) | 1,638 (18.57%)/ 5,009 (14.20%) | 2,005 (22.73%)/ 5,027 (14.25%) | 953 (10.80%)/ 2,030 (5.75%) | 2,314 (26.23%)/ 5,852 (16.59%) | |
| | 530 (6.01%)/ 1,390 (3.94%) | 1,072 (12.15%)/ 2,695 (7.64%) | 1,025 (11.62%)/ 1,760 (4.99%) | 886 (10.04%)/ 1,520 (4.31%) | 1,567 (17.76%)/ 2,831 (8.02%) | |
| | 604 (6.85%)/ 1,981 (5.61%) | 1,560 (17.69%)/ 4,784 (13.56%) | 1,700 (19.27%)/ 3,519 (9.97%) | 800 (9.07%)/ 1,734 (4.91%) | 2,108 (23.90%)/ 4,839 (13.71%) | |
| | - | 1,465 (16.61%)/ 4,311 (12.22%) | 2,041 (23.14%)/ 5,083 (14.41%) | - | | 2,041 (23.14%)/ 5,083 (14.41%) |
| | - | 4,838 (54.85%)/ 17,965 (50.92%) | 5,181 (58.73%)/ 18,175 (51.51%) | - | | 5,181 (58.73%)/ 18,175 (51.51%) |
| | - | 291 (3.30%)/ 910 (2.58%) | 411 (4.66%)/ 942 (2.67%) | - | | 411 (4.66%)/ 942 (2.67%) |
| | 1,198 (13.58%)/ 3,808 (10.79%) | 1,198 (13.58%)/ 3,808 (10.79%) | | 1,670 (18.93%)/ 3749 (10.63%) | 1,670 (18.93%)/ 3749 (10.63%) | |

| SDOH | NLP-extracted Case(%)/Control(%) |
|---|---|
| Social Problems[a] | 2,584 (29.29%) / 7,891 (22.36%) |
| Financial Problems[b] | 2,016 (22.85%) / 6,052 (17.15%) |
| Housing Instability | 1,399 (15.86%) / 4,266 (12.09%) |
| Legal Problems | 730 (8.28%) / 1,644 (4.66%) |
| Violence | 1,141 (12.94%) / 3,228 (9.15%) |
| Barriers to Care | 1,465 (16.61%) / 4,311 (12.22%) |
| Transitions of Care | 4,838 (54.85%) / 17,965 (50.92%) |
| Food Insecurity | 291 (3.30%) / 910 (2.58%) |
| Non-specific Psychosocial Needs | - |

[a]social problems = social or familial problems (Structured) + social isolation (NLP)
[b]financial problems = employment or financial problems (Structured) + job or financial insecurity (NLP)

**Table 2.** Summary Statistics of SDOH Factors as Covariate and as Exposure.

All 8 NLP-extracted SDOH were significantly associated with increased risk of death by suicide (Table 3). 'Legal problems' had the largest estimated effect size (more than twice the risk of those with no exposure; aOR = 2.62, 95% CI = 2.38-2.89), followed by 'violence' (aOR = 2.34, 95% CI = 2.17-2.52) and 'social isolation' (aOR = 1.94, 95% CI = 1.83-2.06). All 7 structured SDOH also showed significant associations; again, 'legal problems' had the highest aOR (2.63, 95% CI = 2.37-2,91). Similarly, all combined SDOH showed strong associations and the top three risk factors were 'legal problems' (aOR = 2.66, 95% CI = 2.46-2.89), 'violence' (aOR = 2.12, 95% CI = 1.98-2.27) and 'non-specific psychosocial needs' (aOR = 2.07, 95% CI = 1.92-2.23).

| SDOH factors | NLP-extracted, aOR (95% CI)[a] | Structured, aOR (95% CI)[a] | Combined, aOR (95% CI)[a] |
|---|---|---|---|
| Social problems[b] | 1.94 (1.83, 2.06) | 2.11 (1.94, 2.29) | 1.95 (1.84, 2.07) |
| Financial problems[c] | 1.91 (1.79, 2.04) | 2.18 (1.97, 2.42) | 1.92 (1.80, 2.05) |
| Housing instability | 1.90 (1.78, 2.03) | 2.28 (2.06, 2.53) | 1.93 (1.80, 2.06) |
| Legal problems | 2.62 (2.38, 2.89) | 2.63 (2.37, 2.91) | 2.66 (2.46, 2.89) |
| Violence | 2.34 (2.17, 2.52) | 1.96 (1.77, 2.16) | 2.12 (1.98, 2.27) |
| Barriers to care | 1.86 (1.74, 1.99) | - | 1.86 (1.74, 1.98) |
| Transition of care | 1.53 (1.44, 1.62) | - | 1.51 (1.43, 1.60) |
| Food insecurity | 1.85 (1.62, 2.11) | - | 1.85 (1.62, 2.11) |
| Non-specific psychosocial needs | - | 2.09 (1.94, 2.25) | 2.07 (1.92, 2.23) |

[a]Each model was adjusted for socio-demographic variables, psychiatric symptoms, substance abuse, pain, patient disability, clinical comorbidities and all SDOH in its group.
[b]social problems = social or familial problems (Structured) + social isolation (NLP)
[c]financial problems = employment or financial problems (Structured) + job or financial insecurity (NLP)

**Table 3.** Associations of SDOH with Veterans' death by suicide

When considered simultaneous exposure to two SDOH, we found all combinations of SDOH to be strongly associated with increased risk of death by suicide (**Appendix 5**), regardless of the SDOH extraction process. For NLP-extracted SDOH, the highest aOR was for exposure to 'legal problems' and 'violence' (aOR = 3.44, 95% CI = 3.03-3.89). For structured SDOH, exposure to 'financial problems' and 'violence' had the highest aOR (3.54 95% CI = 2.87-4.36). Combined SDOH also showed a similar trend.

**Discussion**

To our knowledge, this is the first large-scale study that used both structured and unstructured EHR data to investigate the relation between Veterans' suicide and SDOH. We developed and deployed an NLP system to extract SDOH from unstructured clinical notes and found that all NLP-extracted SDOH were strongly associated with increased odds of suicide. We observed similar results for structured and combined SDOH.

Though many studies have explored the effect of various SDOH over different clinical outcomes[14,29–31], very few have examined the association of SDOH with increased risk of suicide, or the magnitude of such association, if any. In a nested case-control study of Veterans, Kim et al.[8] used chart review to examine the impact of SDOH. However, their study focused on a high-risk population of those with depression and had a small sample size (n=636). In contrast, in a large cross-sectional study of Veterans, Blosnich et al.[6] found a dose-response-like association with SDOH for both suicidal ideation and attempt. However, cross-sectional studies are unsuitable for investigating rare events such as suicide [32]. Most importantly, neither of these studies used the rich information provided by clinical notes. On the other hand, in a case-control study, Dobscha et al.[33] extracted SDOH from clinical notes through manual record review and found no evidence of association between Veterans' suicide and SDOH. They had a relatively small sample size (n=783) and included only male patients.

An important contribution of our study is the development of an NLP system to extract SDOH from unstructured EHR text. Our NLP system extracted a considerable number of SDOH that were not available from the structured data fields (**Appendix 4**). These can help providers identify crucial SDOH information that they would otherwise miss. However, NLP-extracted SDOH did not cover all structured

SDOH. Across the 5 common SDOH, NLP extracted 44.91% of the structured SDOH information as covariates whereas as exposures it extracted 49.92%. This may be due to missing SDOH information in EHR notes or false negatives from the NLP system. Structured data, on the other hand, identified 18.86% of the NLP-extracted SDOH as covariates and 22.85% as exposures. Therefore, taking their unique contributions into account, we suggest combining both structured SDOH and NLP-extracted SDOH for assessment.

For the 5 common SDOH, structured SDOH consistently showed higher aORs for suicide than NLP-extracted SDOH. One possible explanation for this might be that in controls, who are less likely to be sick, clinicians may not be inclined to note their SDOH information in the structured data fields. We hypothesize that clinicians only do so when it is pertaining to the patient's primary diagnosis or ongoing clinical care, possibly representing a relatively sicker population than all the patients with identifiable SDOH in their clinical notes. For example, 14.64% of the case population were exposed to social problems, as identified by the structured data, compared with 34.92% by the NLP system - a 2.4 times increase (Table 2). However, this goes up to 2.8 times for the controls (7.96% vs 22.13%). Thus, using NLP-derived SDOH information might reduce information bias, an important problem in assessing psychosocial research questions.

To estimate whether intervening on SDOH has the potential to change suicide risk, it is necessary to separate its influence from other related factors. In effect, we aimed at emulating the results of an experimental setting where people who experience certain SDOH issues would be enrolled in a trial that randomly assigns whether one receives an intervention. Because such a trial is not available, we relied on observational health data to inform our understanding of suicide. We used epidemiologic methods to adjust for the differences between people exposed to SDOH and those who were not. We carefully considered several possible confounding health and demographic factors in our design to obtain the best possible estimate of the associations of SDOH on suicide.

Our work shows a strong impact of SDOH on Veterans' risk of suicide using a nested case-control design, in which both the covariate and exposure assessment periods are limited to two years. This setup reduces the burden of data processing and NLP extraction, and yet provides a valid assessment of the potential associations between (recent) SDOH and suicide. On the other hand, using longer covariate and exposure assessment periods could provide more information and insights on both short-term (acute) and long-term (persistent) impacts of SDOH on suicide. A related problem is that SDOH change over time; as such, it is more appropriate to treat them as time-varying exposures for longer exposure assessment periods. These time-varying aspects of the problem will be carefully explored in our future work.

**Limitations**
Our study has some limitations. First, the VA population does not represent the general US population. However, many studies and innovations from the VHA have been shown to assist non-VHA facilities in adopting better clinical practices[34–36]. Second, there is potential for residual confounding. Third, EHR data might have incomplete or missing SDOH information[37], making it challenging to assess the influence of SDOH on any target outcome. However, most SDOH with a direct relation to provided care are recorded, so our approach is unlikely to miss important SDOH when both structured and unstructured data are used.

**Conclusions**

Ours is the first large-scale study to implement and use an NLP system to extract SDOH information from unstructured EHR data. We showed that SDOH can significantly contribute to Veterans' death by suicide. Our results also indicate that integrating NLP-based SDOH can benefit similar analyses by identifying more patients at risk. We strongly believe that analyzing all available SDOH information, including those contained in clinical notes, can help develop a better system for risk assessment and suicide prevention. However, more studies are required to investigate ways of seamlessly incorporating SDOH into existing healthcare systems.

**Acknowledgements**

We thank our annotators Raelene Goodwin, Heather Keating, and Emily Druhl for annotating EHR notes that were essential for training our NLP system. A Mitra and H Yu had full access to all the data in the study and takes responsibility for the integrity of the data and the accuracy of the data analysis This work was funded by the grant R01MH125027 from the National Institute of Mental Health (NIMH) of the National Institutes of Health (NIH). The funding source had no role in the design and conduct of the study; collection, management, analysis, and interpretation of the data; preparation, review, or approval of the manuscript; and decision to submit the manuscript for publication. The contents of this paper do not represent the views of NIH, VA, or the United States Government.

**Appendix**
**Appendix 1: NLP Model Development**
**Dataset**
We used a stratified random sampling approach to sample 3,000 Veterans who received treatment at the VHA and died between fiscal year 2009-2017. Patients were stratified and sampled by several key sociodemographic variables (race, gender, and age), geographic location (Northeast, Midwest, South, and West, 1:1:1:1 ratio), and death by suicide (2:1 ratio to those who did not die from suicide). We also oversampled underrepresented groups such as women and ethnic minorities. For the ease of manual annotation, we further reduced the number of Veterans by sampling 1,393 veterans while maintaining a Veterans with suicide attempt to no attempt ratio of 1:4. For each Veteran, we chose three types of notes – social worker notes, mental health notes and emergency room visit notes.

At first, we conducted a pre-screening phase using a predefined set of keywords to select the most relevant sentences in a note and turned each of them into a three-sentence paragraph by using the previous and next sentences. All these paragraphs came from 4,646 unique EHR notes. Next, these paragraphs were annotated for 13 distinct SDOH, behavioral and other relevant factors by three expert annotators under expert supervision (Table 1). These 13 factors were selected based on expert opinions and the recent clinical practice guideline issued by the Veterans Affairs and Department of Defense [1]. Each factor was further annotated for two attributes, 'presence' (yes, not yes) and 'period' (current, not current).

| NLP Extracted Variables | Brief descriptions | Example tokens |
|---|---|---|
| Social isolation* | Social and behavioral status to detect loneliness, lack of social and/or family support; marital/relationship status. | Alone, lonely, divorce, widow etc. |
| Transition of care* | Change of admission status (discharge, transfer etc.); change in medication and/or provider. | Discharge, admission, change in medication, transfer etc. |
| Barriers to care* | Transportation issues; communication problems; lack of trust or rapport; intellectual disability | Transportation issues, garbled speech, communication problems etc. |
| Financial insecurity* | Financial issues; job problems; poverty | Unemployed, poor, unemployment, rehabilitation etc. |
| Housing instability* | Housing issues | Eviction, homeless, homelessness etc. |
| Food insecurity* | Poor diet and/or nutrition; lack of access to proper meal; dependency on food charities/voucher/stamps | Hungry, pantry, starvation, food voucher etc. |
| Violence* | Availability and/or access to lethal means; bullying; domestic violence; any harassment/abuse/trauma; racism; homicidal ideation; feeling scared/unsafe | Firearms, violence, assault, weapon, abuse, homicidal, racism etc. |
| Legal problems* | Imprisonment; court-related matters; detentions; disciplinary action; restraining orders; brushes with the law; criminal charges; any violation of law | Imprisonment, parole, arrested, felony, investigation, prison etc. |
| Substance abuse | Drug use disorder; alcohol consumption; alcohol use disorder; addiction; overdose | Alcohol, tobacco, heroin, cocaine, smoking, overdose etc. |
| Psychiatric symptoms | Hopelessness; Insomnia; Problem solving difficulty; decreased | PTSD, depression, anxiety, schizophrenia, insomnia, hallucination |

| | psychosocial functioning; psychiatric hospitalization; eating disorder; mention of any psychiatric disease | etc. |
|---|---|---|
| Pain | Physical pain | Pain, suffering, hurting, discomfort etc. |
| Patient disability | Reliance on assistive devices and/or disability pay; service-connected ratings | Disabled, blind, hearing loss, wheelchair etc. |
| Suicide outcome | Suicide attempt and/or ideation | Feel like shooting myself, no desire to live, better off dead |

*SDOH

**eTable 1.** Examples of NLP-extracted factors.

**Task and Experimental Setup**

We designed this as a sequence labeling problem where, given a sentence from a note, our goal was to label each word with the appropriate SDOH/behavioral factor and detect the two attributes – presence and period. To be specific, we implemented a multi-task learning framework. A 60:20:20 split was used for the train, validation, and test sets. We experimented with four pretrained language models, namely, RoBERTa [2], BioBERT [3], Bio+Clinical BERT [4] and EhrBERT [5], and for each, we added three heads on top, to identify factors and their attributes jointly. Our code is publicly available at https://github.com/avipartho/SequenceLabelingWithMultiTaskLEarning.

**Result**

We evaluated the model performance using precision, recall and F-score. We considered both exact matching (i.e., both the span boundary and entity type match with that of the ground truth) and relaxed matching (i.e., the span boundaries overlap and entity types match) to assess the model performance. In our experiments, RoBERTa achieved the best performance among all the models. The results for RoBERTa are shown in eTable 2, eTable 3 and eTable 4.

| SDOH/behavioral Factors | Exact | | | Relaxed | | |
|---|---|---|---|---|---|---|
| | Precision | Recall | F-score | Precision | Recall | F-score |
| Social isolation* | 85.45 | 88.04 | 86.73 | 92.05 | 94.84 | 93.42 |
| Transition of care* | 80.07 | 91.92 | 85.59 | 81.21 | 93.23 | 86.81 |
| Barriers to care* | 54.66 | 62.93 | 58.50 | 68.43 | 78.78 | 73.24 |
| Financial insecurity* | 64.52 | 76.04 | 69.81 | 74.43 | 87.71 | 80.53 |
| Housing instability* | 74.37 | 82.24 | 78.11 | 81.06 | 89.64 | 85.13 |
| Food insecurity* | 66.87 | 76.43 | 71.33 | 70.63 | 80.71 | 75.33 |
| Violence* | 67.01 | 74.71 | 70.65 | 79.71 | 88.88 | 84.04 |
| Legal problems* | 64.45 | 68.86 | 66.58 | 82.35 | 87.98 | 85.07 |
| Substance abuse | 70.75 | 78.61 | 74.48 | 82.32 | 91.47 | 86.65 |
| Psychiatric symptoms | 74.72 | 82.64 | 78.48 | 82.46 | 91.20 | 86.61 |
| Pain | 77.82 | 85.88 | 81.65 | 86.86 | 95.86 | 91.14 |
| Patient disability | 85.91 | 89.35 | 87.60 | 90.41 | 94.03 | 92.19 |
| Suicide outcome | 67.43 | 73.89 | 70.51 | 81.12 | 88.89 | 84.82 |
| Micro | 75.03 | 82.36 | 78.52 | 83.14 | 91.27 | 87.02 |
| Macro | 71.85 | 79.35 | 75.39 | 81.00 | 89.48 | 85.00 |

eTable 2. Performance of our MTL model – factor identification.

| Presence | Exact | | | Relaxed | | |
|---|---|---|---|---|---|---|
| | Precision | Recall | F-score | Precision | Recall | F-score |
| Yes | 71.86 | 77.05 | 74.36 | 81.20 | 87.06 | 84.02 |
| Not yes | 74.14 | 76.43 | 75.27 | 79.76 | 82.22 | 80.97 |
| Micro | 72.68 | 76.82 | 74.69 | 80.68 | 85.28 | 82.92 |
| Macro | 73.00 | 76.74 | 74.82 | 80.48 | 84.64 | 82.50 |

eTable 3. Performance of our MTL model – presence identification.

| Presence | Exact | | | Relaxed | | |
|---|---|---|---|---|---|---|
| | Precision | Recall | F-score | Precision | Recall | F-score |
| Current | 75.65 | 79.84 | 77.69 | 83.78 | 88.42 | 86.04 |
| Not current | 58.93 | 64.94 | 61.79 | 68.28 | 75.24 | 71.59 |
| Micro | 73.69 | 78.16 | 75.86 | 81.96 | 86.94 | 84.38 |
| Macro | 67.29 | 72.39 | 69.74 | 76.03 | 81.83 | 78.81 |

eTable 4. Performance of our MTL model – period identification.

## Appendix 2: ICD and Stop Codes for structured SDOH and Mental Health Disorders

**SDOH**

| SDOH factors | ICD-9 Codes | Stop Codes |
|---|---|---|
| Social or familial problems | V60.6, V60.89, V62.3, V62.89, V61, V62.4 | - |
| Employment or financial problems | V60.2, V60.89, V60.9, V62.0, V62.1, V62.29 | 208, 222, 535, 555, 568, 574 |
| Housing instability | V60.0-2, V60.89 | 504, 507, 508, 511, 522, 528, 529, 530, 555, 556, 590 |
| Legal problems | V62.5, V62.89, E849.7 | 591, 592 |
| Violence | E904.0, E960.0-1, E961-E977, E979, E990.0-3, E990.9, E991.0-9, E992.0-3, E992.8-9, E993.0-9, E994.0-3, E994.8-9, E995.0-4, E995.8-9, E996.0-3, E996.8-9, E997.0-3, E997.8-9, E998.0-1, E998.8-9, E999.0-1, V15.41-42, V15.49, V71.5, V71.81, 995.50-54, 995.80-85 | 524 |

| Non-specific psychosocial needs | V62.29, V62.3-6, V62.81, V62.89, V62.9 | - |

eTable 5. ICD and stop codes for structured SDOH.

**Comorbidities**

| Mental Health Disorders | ICD-9 Codes |
|---|---|
| Major Depressive disorder | 293.83, 296.2, 296.3, 296.9, 298.0, 300.4, 301.12, 309.0, 309.1, and 311 |
| Alcohol Use Disorder | 291.0-5, 291.8-9, 303.0-303.9, 305.0, 357.5, 425.5, 571.0-3, 535.3, V11.3 |
| Drug Use Disorder | 292.0-1, 304.0-304.9, 305.2-305.8 |
| Anxiety Disorder | 300.0, 300.1, 300.2, 799.2 |
| Posttraumatic Stress Disorder | 309.81 |
| Schizophrenia | 295.0-295.9, V11.0 |
| Bipolar disorder | 296.0, 296.1, 296.4-7, 296.80, 296.81, 296.82, 296.89, 296.90, 296.99, V11.1 |

eTable 6. ICD codes for mental health disorders

## Appendix 3: Base Cohort Statistics

|  | Base Cohort (N=6,122,785) | % |
|---|---|---|
| Race | | |
| White | 4,713,683 | 76.99% |
| Black | 997,035 | 16.28% |
| Asian | 58,075 | 0.95% |
| Native Hawaiian or Other Pacific Islander | 50,210 | 0.82% |
| American Indian | 44,028 | 0.72% |
| Unknown | 259,754 | 4.24% |
| Gender | | |

| | | | |
|---|---|---|---|
| | Male | 5,646,838 | 92.23% |
| | Female | 475,947 | 7.77% |
| Age | | | |
| | 18-29 | 403,618 | 6.59% |
| | 30-39 | 435,592 | 7.11% |
| | 40-49 | 640,441 | 10.46% |
| | 50-59 | 1,072,726 | 17.52% |
| | 60-69 | 1,896,202 | 30.97% |
| | 70-79 | 938,691 | 15.33% |
| | 80-100 | 735,515 | 12.01% |
| Marital Status | | | |
| | Married | 1,672,089 | 27.31% |
| | Single | 330,837 | 5.40% |
| | Divorced | 598,341 | 9.77% |
| | Widowed | 200,842 | 3.28% |
| | Unknown | 3,320,676 | 54.23 |

**eTable 7.** Summary Statistics of the Base Cohort

## Appendix 4: SDOH prevalence

| SDOH | NLP only | Structured data only | Present in both |
|---|---|---|---|
| Social problems | 71.03% | 13.89% | 15.08% |
| Financial insecurity | 74.44% | 8.92% | 16.64% |
| Housing insecurity | 66.17% | 14.77% | 19.06% |
| Legal problems | 49.03% | 36.98% | 13.99% |
| Violence | 59.25% | 31.13% | 9.62% |

**eTable 9.** Prevalence of Combined SDOH Factors by Source (as Covariates)

| SDOH | NLP only | Structured data only | Present in both |
|---|---|---|---|
| Social problems | 67.69% | 14.20% | 18.11% |
| Financial insecurity | 69.19% | 10.23% | 20.58% |
| Housing insecurity | 63.47% | 13.89% | 22.64% |
| Legal problems | 45.29% | 36.68% | 18.03% |
| Violence | 63.52% | 24.87% | 11.61% |

**eTable 10.** Prevalence of Combined SDOH Factors by Source (as Exposures)

## Appendix 5: Associations for concurrent SDOH

| SDOH factors | NLP-extracted, aOR (95% CI)* | Structured, aOR (95% CI)* | Combined, aOR (95% CI)* |
|---|---|---|---|
| Social problems, Financial problems | 2.39 (2.22, 2.59) | 2.48 (2.18, 2.82) | 2.34 (2.17, 2.52) |
| Social problems, Housing instability | 2.47 (2.28, 2.68) | 2.41 (2.11, 2.74) | 2.38 (2.20, 2.57) |
| Social problems, Legal problems | 3.01 (2.70, 3.36) | 2.60 (2.33, 2.92) | 2.84 (2.60, 3.09) |
| Social problems, Violence | 2.94 (2.70, 3.21) | 3.37 (2.82, 4.02) | 2.69 (2.49, 2.91) |
| Social problems, Barriers to care | 2.46 (2.27, 2.67) | - | 2.41 (2.23, 2.61) |
| Social problems, Transition of care | 2.15 (2.02, 2.29) | - | 2.12 (2.00, 2.26) |
| Social problems, Food insecurity | 2.19 (1.87, 2.56) | - | 2.15 (1.85, 2.50) |
| Social problems, Non-specific psychosocial needs | - | 2.15 (1.96, 2.36) | 2.21 (2.04, 2.40) |
| Financial problems, Housing instability | 2.45 (2.25, 2.66) | 2.10 (1.86, 2.35) | 2.35 (2.17, 2.55) |
| Financial problems, Legal problems | 3.00 (2.68, 3.35) | 3.16 (2.66, 3.75) | 2.98 (2.70, 3.27) |
| Financial problems, Violence | 2.83 (2.58, 3.11) | 3.54 (2.87, 4.36) | 2.69 (2.47, 2.94) |
| Financial problems, Barriers to care | 2.35 (2.16, 2.57) | - | 2.33 (2.14, 2.54) |
| Financial problems, Transition of care | 2.07 (1.94, 2.22) | - | 2.08 (1.94, 2.22) |
| Financial problems, Food insecurity | 2.19 (1.87, 2.56) | - | 2.20 (1.88, 2.57) |
| Financial problems, Non-specific psychosocial needs | - | 2.45 (2.14, 2.80) | 2.37 (2.16, 2.60) |
| Housing instability, Legal problems | 3.16 (2.81, 3.56) | 2.91 (2.45, 3.45) | 3.14 (2.84, 3.47) |
| Housing instability, Violence | 2.95 (2.67, 3.27) | 3.13 (2.52, 3.88) | 2.77 (2.52, 3.04) |
| Housing instability, Barriers to care | 2.42 (2.20, 2.66) | - | 2.41 (2.20, 2.64) |
| Housing instability, Transition of care | 2.08 (1.93, 2.23) | - | 2.08 (1.93, 2.23) |
| Housing instability, Food insecurity | 2.21 (1.87, 2.61) | - | 2.18 (1.85, 2.56) |
| Housing instability, Non-specific psychosocial needs | - | 2.64 (2.30, 3.03) | 2.46 (2.23, 2.72) |
| Legal problems, Violence | 3.44 (3.03, 3.89) | 3.31 (2.72, 4.02) | 3.34 (3.01, 3.70) |
| Legal problems, Barriers to care | 3.00 (2.65, 3.39) | - | 2.93 (2.63, 3.25) |
| Legal problems, Transition of care | 2.78 (2.51, 3.07) | - | 2.77 (2.54, 3.02) |
| Legal problems, Food insecurity | 2.73 (2.20, 3.39) | - | 2.66 (2.20, 3.23) |
| Legal problems, Non-specific psychosocial needs | - | 2.65 (2.39, 2.94) | 2.74 (2.48, 3.01) |
| Violence, Barriers to care | 2.82 (2.55, 3.12) | - | 2.65 (2.41, 2.91) |
| Violence, Transition of care | 2.51 (2.32, 2.71) | - | 2.31 (2.15, 2.48) |
| Violence, Food insecurity | 2.54 (2.09, 3.09) | - | 2.39 (1.99, 2.88) |
| Violence, Non-specific psychosocial needs | - | 2.93 (2.50, 3.43) | 2.83 (2.55, 3.15) |
| Barriers to care, Transition of care | 2.00 (1.86, 2.14) | - | 1.99 (1.86, 2.13) |
| Barriers to care, Food insecurity | 2.19 (1.86, 2.59) | - | 2.20 (1.87, 2.60) |
| Barriers to care, Non-specific psychosocial needs | - | - | 2.40 (2.16, 2.66) |
| Transition of care, Food insecurity | 1.95 (1.70, 2.24) | - | 1.96 (1.70, 2.25) |
| Transition of care, Non-specific psychosocial needs | - | - | 2.23 (2.05, 2.42) |
| Food insecurity, Non-specific psychosocial needs | - | - | 2.48 (2.04, 3.02) |

*Each model was adjusted for socio-demographic variables, psychiatric symptoms, substance abuse, pain, patient disability, clinical comorbidities and all SDOH in its group.

**eTable 8.** Associations of SDOH with Veterans' death by suicide